\documentclass[runningheads,a4paper]{llncs}
\usepackage[utf8]{inputenc}
\usepackage{amsmath}
\usepackage{amsfonts}
\usepackage{amssymb}
\usepackage{booktabs}
\usepackage[dvips]{hyperref}
\usepackage{graphicx}
\usepackage{algorithmicx}
\usepackage{algpseudocode}
\usepackage{algorithm}
\usepackage{color}
\usepackage{url}
\usepackage{cite}
\usepackage{subfigure}
\usepackage{multirow}
\usepackage[raggedright]{sidecap}
\sidecaptionvpos{figure}{t} 

\def\x{{\mathbf x}}

\def\R{{\mathbf R}}
\def\O{{\mathbf O}}
\def\bS{{\mathbf S}}
\def\post{{\gamma}}
\def\bpost{{\boldsymbol \gamma}}
\def\meanvec{{\boldsymbol \mu}}
\def\covmat{{\boldsymbol \Sigma}}

\def\ndist{{\mathcal{N}}}
\newcommand{\observed}[1]{{$#1^\text{o}$}}
\newcommand{\missing}[1]{{$#1^\text{m}$}}

\def\X{{\mathbf X}}
\def\Xm{{\X^\text{m}}}

\def\Xo{{\X^\text{o}}}

\def\R{{\mathbf R}}
\def\r{{\mathbf r}}
\def\Y{{\mathbf Y}}

\def\PCKID{{\textit{PCKID}}}

\definecolor{darkgreen}{rgb}{0,0.6,0.2}

\newcommand{\algoref}[1]{{Alg.~\ref{#1}}}
\newcommand{\figref}[1]{{Fig.~\ref{#1}}}
\newcommand{\tabref}[1]{{Tab.~\ref{#1}}}
\newcommand{\secref}[1]{{Sec.~\ref{#1}}}

\newcommand{\keywords}[1]{\par\addvspace\baselineskip
\noindent\keywordname\enspace\ignorespaces#1}

\begin{document}

\mainmatter  

\title{Spectral Clustering using \PCKID{} -- A Probabilistic Cluster Kernel for Incomplete Data}

\titlerunning{Spectral Clustering using \PCKID{}}

%
%
\author{Sigurd L\o{}kse\inst{1}\thanks{sigurd.lokse@uit.no} \and Filippo M. Bianchi\inst{1}
    \and Arnt-B\o{}rre Salberg\inst{2} \and Robert Jenssen\inst{1,2}}
\authorrunning{Løkse et al.}

\institute{Machine Learning Group\thanks{\url{http://site.uit.no/ml}},
    UiT -- The Arctic University of Norway
    \and Norwegian Computing Center}

%
%

\toctitle{Spectral clustering using PCKID}
\tocauthor{}
\maketitle

\begin{abstract}
In this paper, we propose \PCKID{}, a novel, robust, kernel function
    for spectral clustering, specifically designed to handle incomplete data.
By combining posterior distributions of Gaussian Mixture Models for
    incomplete data on different scales, we are able to learn a
    kernel for incomplete data that does not depend on any critical
    hyperparameters, unlike the commonly used RBF kernel.
To evaluate our method, we perform experiments on two real datasets.
\PCKID{} outperforms the baseline methods for all fractions of missing values and
    in some cases outperforms the baseline methods with up to 25 percentage points.

\keywords{Missing data, robustness, kernel methods, spectral clustering}
\end{abstract}

\section{Introduction}
Clustering is of utmost importance in the field of machine learning,
    with a huge literature and many practical applications\cite{jain2010data}.
Over the past decades, a huge variety of methods have been proposed.
These range from simple linear methods like $k$-means
    \cite{theodoridis2008pattern}, to more recent
    advanced methods, like spectral clustering
    \cite{ng2001spectral, von2007tutorial, filippone2008survey, yang2010image, nie2011spectral}.
Spectral clustering is a family of highly performing clustering algorithms,
    currently considered state of the art.
In spectral clustering, the eigenvectors and eigenvalues (spectrum) of some
    similarity matrix are exploited to generate a beneficial
    representation of the data, such that a simple method
    like $k$-means could be utilized to generate a partitioning,
    even with non-linearly separable data.

Analyzing incomplete datasets (with missing features) is a big challenge within
    clustering methods and data analysis in general, since encountering incomplete
    data is common in real applications.
For instance, an entry in the dataset may not be recorded if a sensor is failing
    or a field in a questionnaire is left unanswered.
Both supervised and unsupervised methods have been proposed to deal with incomplete data.
In the supervised setting, we have e.g. a max--margin framework, where geometric interpretations of
    the margin is used to account for missing data \cite{chechik2008max},
    an approach based on training one SVM per missingness pattern \cite{salberg2012land} and
    the ''best'' Bayesian classifier \cite{mojirsheibani2007statistical} approach.
In the unsupervised setting, there are mixture model formulations accounting for missing
    features, including both non--Bayesian approaches \cite{ghahramani1994supervised, lin2006fast}
    and Bayesian approaches \cite{marlin2008missing}.
In general, a common approach is to apply imputation techniques \cite{dixon1979pattern} to estimate
    the missing values and then proceeding with the analysis on the imputed, complete, data set.
None of these approaches come without challenges since the best choice of imputation
    technique is often very dependent on the data, and moreover difficult to evaluate. 
    

In this paper, we propose as a new approach to integrate in a synergistic manner
    recent advances in spectral clustering and kernel methods with existing
    probabilistic methods for dealing with incomplete data. 
In particular, we exploit the Probabilistic Cluster Kernel (PCK)
    framework \cite{izquierdo2015spectral}, which combines posterior distributions
    of Gaussian Mixture Models (GMMs) on different scales to learn a robust
    kernel function, capturing similarities on both a global and local scale.
This kernel function is robust with regards to hyperparameter choices, since instead of
    assuming some structure in the data, the ensemble of GMMs adapt to the data manifold. 
We hyptothesize that by integrating GMMs specifically designed to handle incomplete data
    \cite{lin2006fast} into the PCK framework for spectral clustering, we will be
    able to cluster incomplete data sets in a more robust manner compared to existing
    approaches. The proposed approach for building the kernel matrix to be used for
    spectral clustering in our framework, is denoted the \emph{Probabilistic Cluster Kernel for Incomplete Data (\PCKID{})}.
  

\section{Background theory}

\subsection{Missing data mechanisms} \label{sec:missing_mechanisms}
Let $\x = \{x_{i}\}$ denote a data vector and let \observed{\x} and \missing{\x} denote the observed-
  and missing features of $\x$.
Define $\r = \{r_{i}\}$, where $r_{i} = 1$ if $x_{i} \in \;$\missing{\x} and zero otherwise to
  be the \textit{missing indicator} for $\x$.  
In order to train a model that accounts for values in the dataset that are not observed,
    one has to rely on  assumptions that describe how missing data occurs.
In this section, we describe the three main missing data mechanisms that characterize
    the structure of $\r$ \cite{salberg2012land}. 

\subsubsection{Missing completely at random (MCAR)}
Features are said to be \textit{missing completely at random} (MCAR) if the features are missing
  independently from both the observed values \observed{\x} and the missing values \missing{\x}.
That is,
  \[
    P(\R | \X) = P(\R).
  \]
This is the missingness assumption on the data that leads to the simplest analysis.
However, this assumption is rarely satisfied in practice.

\subsubsection{Missing at random (MAR)}
If the \textit{features} are missing independently of their \textit{values}, the features are said
  to be \textit{missing at random} (MAR).
Then the missingness of the features are only dependent of the \textit{observed} values,
  such that
\[
  P(\R | \X) = P\left(\R | \Xo\right).
\]
This missing data mechanism is often assumed when working with missing data, since
  many real world missing data are generated by this mechanism.
For instance, a blood test of a patient might be missing if it is only taken given some other
  test (observed value) exceeds a certain value.

\subsubsection{Not missing at random (NMAR)}
If the missingness of a feature is dependent on their values, it is said to be not missing at random
  (NMAR), that is
\[
  P(\R | \X) = P\left(\R | \Xm \right).
\]
For instance, NMAR occurs when a sensor measurement is discarded because it goes beyond the maximum
  value that the sensor can handle.

\subsection{Gaussian Mixture Models for Incomplete Data}
In this section, we briefly summarize how to implement Gaussian Mixture Models (GMM)
    when the data have missing features.
This model will be exploited as the foundation for \PCKID{} to learn a robust kernel
    function.
For details, we address the interested reader to \cite{lin2006fast}.

A GMM is used to model the probability density function (PDF) for given dataset.
In a GMM, a data point $\x_i$ is assumed to be sampled from a multivariate
    Gaussian distribution
    $\ndist_k(\mathbf x_i | \meanvec_k, \covmat_k)$ with probability $\pi_k$ and
    $k \in [1,K]$, where $K$ corresponds to the number of mixture components.
Accordingly, the PDF of the data is modeled by a \textit{mixture} of Gaussians, such that
\begin{equation}
    \label{eq:mixture_model}
    f(\x) = \sum_{k = 1}^K \pi_k \ndist(\x | \meanvec_k, \covmat_k).
\end{equation}
The maximum likelihood estimates for the parameters in this model
    can be approximated through the Expectation Maximization (EM) algorithm.
    
When the data have missing features, we assume that the elements in a data vector $\x_i$
    can be partitioned into two components; one observed part \observed{\x_i} and one
    missing part \missing{\x_i} as explained in \secref{sec:missing_mechanisms}.
Then, one can construct a binary matrix $\O_i$ by removing the rows from the identity matrix corresponding to the missing elements \missing{\x_i}, such that \observed{\x_i}$\; = \O_i\x_i$.
Given the mean vector $\meanvec_k$ and the covariance matrix $\covmat_k$ for mixture component $k$,
    the mean and covariance matrix for the \textit{observed} part of missingness pattern $i$
    is given by 
\begin{equation*}
    \begin{aligned}
    \meanvec_{k, i}^\text{o} &= \O_i\meanvec_k \\
    \covmat_{k, i}^\text{o} &= \O_i \covmat_k \O_i^T.
    \end{aligned}
\end{equation*}
By defining
\begin{equation*}
    \bS_{k, i}^\text{o} = \O_i^T {\covmat_{k, i}^\text{o}}^{-1} \O_i,
\end{equation*}
    one can show that, under the MAR assumption, the EM procedure outlined in  
    \algoref{alg:emincompletegmm} will find the parameters that maximizes
    the likelihood function \cite{lin2006fast}.

Note that, even though the notation in this paper allows for a unique missingness
    pattern for each data point $\x_i$, one missingness pattern is usually
    shared between several data points.
Thus, to improve efficiency when implementing \algoref{alg:emincompletegmm}, one
    should sort the data points by missingness pattern such that parameters that
    are common across data points are calculated only once \cite{lin2006fast}. 
    
\begin{algorithm}[tbp] \scriptsize
    \caption{EM algorithm for incomplete data GMM}
    \label{alg:emincompletegmm}
    \begin{algorithmic}[1]
    \State{Initialize $\hat\meanvec_k^{(0)}$, $\hat\covmat_k^{(0)}$, $\hat\pi_k^{(0)}$ and $\hat\post_{i, k}^{(0)}$ for $k \in [1, K]$ and $i \in [1, N]$.}
    \While{not converged}
        \State{\textbf{E-Step:}} Compute
        \begin{equation*}
            \begin{aligned}
            \hat{\post}_{k, i}^{(\ell)} &= \frac{
                    \hat\pi_k^{(\ell)}
                    \mathcal{N}\left(
                        \x_i^\text{o} |
                        \hat\meanvec_{k, i}^{\text{o}(\ell)},
                        \hat\covmat_{k, i}^{\text{o}(\ell)}
                    \right)
                }{
                    \sum_{j = 1}^K  
                    \hat\pi_j^{(\ell)}
                    \mathcal{N}\left(
                        \x_i^\text{o} |
                        \hat\meanvec_{j, i}^{\text{o}(\ell)},
                        \hat\covmat_{j, i}^{\text{o}(\ell)}
                    \right)
                }   \\
            \hat{\Y}_{k, i}^{(\ell)} &= \hat\meanvec_k^{(\ell)} + 
                \hat\covmat_k^{(\ell)} \hat\bS_{k, i}^{\text{o}(\ell)}
                \left(\x_i - \hat\meanvec_k^{(\ell)}\right)
            \end{aligned}
        \end{equation*}
        \State{\textbf{M-Step:}} Compute the next model parameters, given by
        \begin{equation*}
            \begin{aligned}
            \hat\pi_k^{(\ell + 1)} &= \frac{1}{N} \sum_{i = 1}^N \hat\post_{k, i}^{(\ell)} \\
            \hat\meanvec_k^{(\ell + 1)} &=
                \frac{
                    \sum_{i = 1}^N \hat\post_{k, i}^{(\ell)} \hat{\Y}_{k, i}^{(\ell)}
                }{
                    \sum_{i = 1}^N \hat\post_{k, i}^{(\ell)}
                } \\
            \hat\covmat_k^{(\ell + 1)} &=
                \frac{
                    \sum_{i = 1}^N \hat{\boldsymbol \Omega}_{k, i}^{(\ell)}
                }{
                    \sum_{i = 1}^N \hat\post_{k, i}^{(\ell)},
                }
            \end{aligned}
        \end{equation*}
        where
\begin{equation*}
    \begin{aligned}
    \boldsymbol\Omega_{k, i}^{(\ell)} &= \hat\post_{k, i}^{(\ell)}
        \Bigg( 
            \left( 
                \hat\Y_{k, i}^{(\ell)} -  \hat\meanvec_k^{(\ell + 1)}
            \right) 
            \left( 
                \hat\Y_{k, i}^{(\ell)} -  \hat\meanvec_k^{(\ell + 1)}
            \right)^T \\
            &+
            \left(
                \mathbf{I} - \hat\covmat_k^{(\ell)} \hat\bS_{k, i}^{o(\ell)}
            \right)
            \hat\covmat_k^{(\ell)}
        \Bigg).
    \end{aligned}
\end{equation*}
    \EndWhile
    \end{algorithmic}
\end{algorithm}
    
\subsubsection{Diagonal covariance structure assumption.}
In some cases, when the dimensionality of the data is large compared to the number of
    data points, in combination with many missingness patterns, one could consider
    assuming a diagonal covariance structure for the GMM for computational efficiency
    and numerical stability when inverting covariance matrices.
This will of course limit the models to not encode correlations between dimensions, but
    for some tasks it provides a good approximation that is a viable compromise when
    limited computational resources are available.
In this case, covariance matrices are encoded in $d$-dimensional vectors,
    which simplifies the operations in \algoref{alg:emincompletegmm}.

Let $\widehat{\boldsymbol{\sigma}}_k$ be the vector of variances for mixture
    component $k$ and let $\widehat{\mathbf{s}}_{k,i}$ be a vector
    with elements $\widehat{s}_{k,i}(\ell) = \frac{1}{\sigma_k(\ell)}$ if
    element $\ell$ of data point $\mathbf x_i$ is observed and
    $\widehat{s}_{k,i}(\ell) = 0$ otherwise.
Define
\begin{equation}
    \widehat{\mathbf y}_{k, i} =
    \widehat{\boldsymbol \mu}_k +
    \widehat{\boldsymbol \sigma}_k \odot
    \widehat{\mathbf{s}}_{k,i}\odot
    (\mathbf x_i - \widehat{\boldsymbol \mu}_k),
\end{equation}
and
\begin{equation}
    \boldsymbol \omega_{k,i} = \hat\post_{k, i}
    \left(
        (\widehat{\mathbf y}_{k, i} - \widehat{\boldsymbol \mu}_k)
        \odot
        (\widehat{\mathbf y}_{k, i} - \widehat{\boldsymbol \mu}_k)
        + 
        \widehat{\boldsymbol \sigma}_k
        -
        \widehat{\boldsymbol \sigma}_k
        \odot
        \widehat{\mathbf{s}}_{k,i}
        \odot
        \widehat{\boldsymbol \sigma}_k
    \right)
\end{equation}
   where $\odot$ denotes the Hadamard (element wise) product. 
Estimating the parameters with an assumption of diagonal covariance
    structure is then a matter of exchanging $\widehat{\mathbf Y}_{k,i}$ and
    $\boldsymbol{\Omega}_{k,i}$ with
    $\widehat{\mathbf y}_{k,i}$ and $\boldsymbol \omega_{k,i}$ respectively
    in \algoref{alg:emincompletegmm}.
    
\subsection{Spectral clustering using Kernel PCA} \label{sec:spectralclustering}
Spectral clustering is a family of clustering algorithms, where the spectrum, i.e.
    the eigenvalues and eigenvectors, of some similarity matrix is exploited
    for clustering of data separated by non-linear structures \cite{ng2001spectral, von2007tutorial, filippone2008survey, yang2010image, nie2011spectral}.
Most spectral clustering algorithms employ a two-stage approach, with i) a non-linear
    feature generation step using the spectrum and ii) clustering by $k$-means on top of the generated features
    \cite{ng2001spectral, shi2000normalized}.
Some have employed a strategy where the final clustering step
    is replaced by spectral rotations \cite{stella2003multiclass, nie2011spectral}
    or by replacing both steps with kernel $k$-means \cite{dhillon2004kernel}, which 
    is difficult to initialize.
In this work, we employ the two stage approach where we use kernel PCA \cite{scholkopf1997kernel}
    to generate $k$-dimensional feature vectors, for then to cluster these using $k$-means.

\subsubsection{Kernel PCA}
Kernel PCA implicitly performs PCA in some reproducing kernel Hilbert space $\mathcal{H}$
    given a positive semidefinite kernel function
    $\kappa\colon \mathcal{X} \times \mathcal {X} \to \mathbb{R}$,
    which computes inner products in $\mathcal{H}$.
If we define a \textit{kernel matrix}, $\mathbf K$, whose elements are the inner products
    $\kappa(\mathbf x_i, \mathbf x_j) = \langle \phi(\mathbf x_i), \phi(\mathbf x_j)\rangle_\mathcal{H}$,
    this matrix is positive semidefinite, and may be decomposed as
    $\mathbf K = \mathbf E \boldsymbol \Lambda \mathbf E^T$, where
    $\mathbf E$ is a matrix with the eigenvectors as columns and $\boldsymbol \Lambda$ is the
    diagonal eigenvalue matrix of $\mathbf K$.
Then it can be shown that the $k$-dimensional projections onto the principal
    components in $\mathcal{H}$ is given by
\begin{equation}\label{eq:kernel_pca}
    \mathbf Z = \mathbf E_k \boldsymbol \Lambda_k^{\frac{1}{2}},
\end{equation}
    where $\boldsymbol \Lambda_k$ consists of the $k$ largest eigenvalues of $\mathbf K$
    and $\mathbf E_k$ consists of the corresponding eigenvectors.
    
The traditional choice of kernel function is an RBF kernel, defined as
\begin{equation} \label{eq:rbfkernel}
    \kappa(\x_i, \x_j) =
        e^{- \frac{1}{2\sigma^2} \|\x_i - \x_j\|^2},
\end{equation}
    where the $\sigma$ parameter defines the width of the kernel.

\clearpage
\section{\PCKID{} -- A Probabilistic Cluster Kernel for Incomplete Data}
In this paper, we propose a novel procedure to construct a kernel matrix based on models
    learned from data with missing features, which we refer to as \PCKID{}.
In particular, we propose to learn similarities between data points in an unsupervised
    fashion by fitting GMMs to the data with different initial conditions $q \in [1, Q]$
    and a range of mixture components, $g \in [2, G]$ and combine the results
    using the posterior probabilities for the data points.
That is, we define the kernel function as
\begin{equation} \label{eq:pckincomplete}
    \kappa_\PCKID(\x_i, \x_j) = \frac{1}{Z} \sum_{q = 1}^Q \sum_{g = 2}^G \bpost_i^T(q, g) \bpost_j(q, g),
\end{equation}
    where $\bpost_i(q, g)$ is the posterior distribution
    for data point $\x_i$ under the model with initial condition $q$ and $g$ mixture components
    and $Z$ is a normalizing constant.
By using \algoref{alg:emincompletegmm} to train the models, we are able to learn the 
    kernel function from the inherent structures of the data, even when dealing with
    missing features.
In this work, we use this kernel for spectral clustering.

The \PCKID{} is able to capture similarities on both a local and a global scale.
When a GMM is trained with many mixture components, each mixture component covers a small,
    \textit{local} region in feature space.
On the contrary, when the GMM is trained with a small number of mixture components,
    each mixture component covers a large, \textit{global} region in feature space.
Thus, if two data points are similar under models on all scales, they are likely to
    be similar, and will have a large value in the \PCKID{}.
This procedure of fitting models to the data on different scales, ensures robustness
    with respect to parameters, as long as $Q$ and $G$ are set sufficiently large.
Thus, we are able to construct a kernel function that is robust with regards to 
    parameter choice.
This way of constructing a robust kernel is similar to the methodology used in ensemble clustering
    and recent work in spectral clustering \cite{izquierdo2015spectral}.
However, such recent methods are not able to explicitly handle missing data.
    
According to the ensemble learning methodology \cite{monti2003consensus,zimek2013subsampling}, 
    we build a powerful learner by combining multiple weak learners.
Therefore, one does not need to run the EM algorithm until convergence, but instead perform 
    just a few iterations\footnote{For instance, 10 iterations.}.
This also has the positive side-effect of encouraging diversity, providing efficiency and 
    preventing overfitting.
To further enforce diversity, it is beneficial to use sub-sampling techniques to train different
    models on different subsets of the data and evaluate the complete kernel on the full dataset.

\subsection{Initialization}
For each mixture model that is trained, one needs to provide an initialization.
Since we are fitting large models to data that in practice does not necessarily fit these models,
    the initialization needs to be reasonable in order to avoid computational issues
    when inverting covariance matrices.
An initialization procedure that has been validated empirically for the \PCKID{} is
\begin{enumerate}
    \item Use mean imputation to impute missing values.
    \item Draw $K$ random data points from the input data and use them as initial cluster centers.
    \item Run \textit{one} $k$-means iteration to get initial cluster assignments and means.
    \item Calculate the empirical covariance matrix from each cluster and calculate empirical prior probabilities for the mixture model based on the cluster assignments.
\end{enumerate}
Data with imputed values is only used to be able to calculate initial means and covariances.
\emph{When training the model, data without imputed values is used.}

\section{Experiments}
\subsection{Experiment setup}
\subsubsection{\PCKID{} parameters}
In order to illustrate that \PCKID{} does not need any parameter tuning,
    the parameters are set to $Q = G = 30$ for all experiments.
In order to increase diversity, each model in the ensemble is trained on a random 
    subset of $50\%$ of the whole dataset.
The kernel is evaluated on the full dataset, once the models are trained.
Each GMM is trained for 10 iterations with a diagonal covariance structure assumption.

\subsubsection{Baseline methods}
For the baseline methods, missing data is handled with imputation techniques,
    in particular, i) zero imputation, ii) mean imputation iii)
    median imputation and (iv) most frequent value imputation.
To produce a clustering result, each of these imputation techniques is coupled with 
    i) $k$-means on the data and ii) spectral clustering using an RBF kernel,
    where the kernel function is calculated by \eqref{eq:rbfkernel}.

Since no hyperparameters need to be tuned in in \PCKID{}, the kernel width $\sigma$ 
    of the RBF is calculated with a rule of thumb.
In particular, $\sigma$ is set to 20\% of the median pairwise distances
    in the dataset, as suggested in \cite{jenssen2010kernel}.
This is in agreement with unsupervised approaches, where labels
    are not known and cross validation on hyperparameters is not possible.
    
\subsubsection{Performance metric}
In order to assess the performance of \PCKID{}, its supervised clustering accuracy
    is compared with all baseline models.
The supervised clustering accuracy is computed by
\begin{equation}
    ACC = \max_\mathcal{M} \frac{\sum_{i = 1}^n \delta\{ y_i = \mathcal{M}(\widehat y_i)\}}{n},
\end{equation}
    where $y_i$ is the ground truth label, $\widehat y_i$ is the cluster label assigned to
    data point $i$ and $\mathcal{M}(\cdot)$ is the label mapping function that maximizes the
    matching of the labels.
This is computed using the Hungarian algorithm \cite{kuhn1955hungarian}.

\subsubsection{Clustering setup}
Spectral clustering with $k$ clusters is performed by mapping the data
    to a $k$ dimensional empirical kernel space and clustering them
    with $k$-means as described in \secref{sec:spectralclustering}. 
For all methods, $k$-means is run 100 times.
The final clustering is chosen by evaluating the $k$-means cost function
    and choosing the partitioning with the lowest cost.
The number of clusters, $k$, is assumed known.

\subsection{MNIST 5 vs. 6}
\begin{figure}[tbp]
    \centering
    \subfigure[]{
        \includegraphics[width=0.12\textwidth]{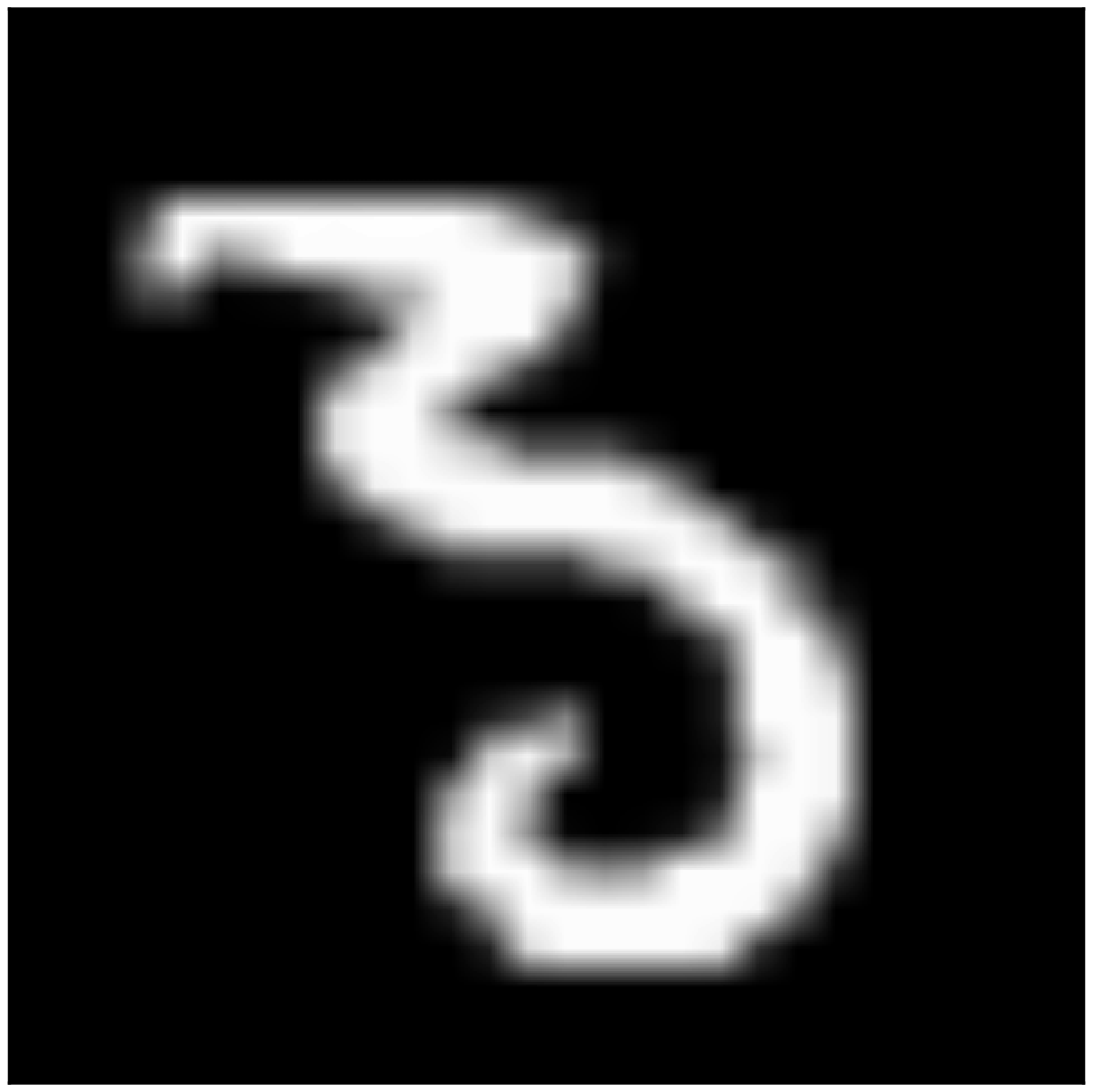}
        \includegraphics[width=0.12\textwidth]{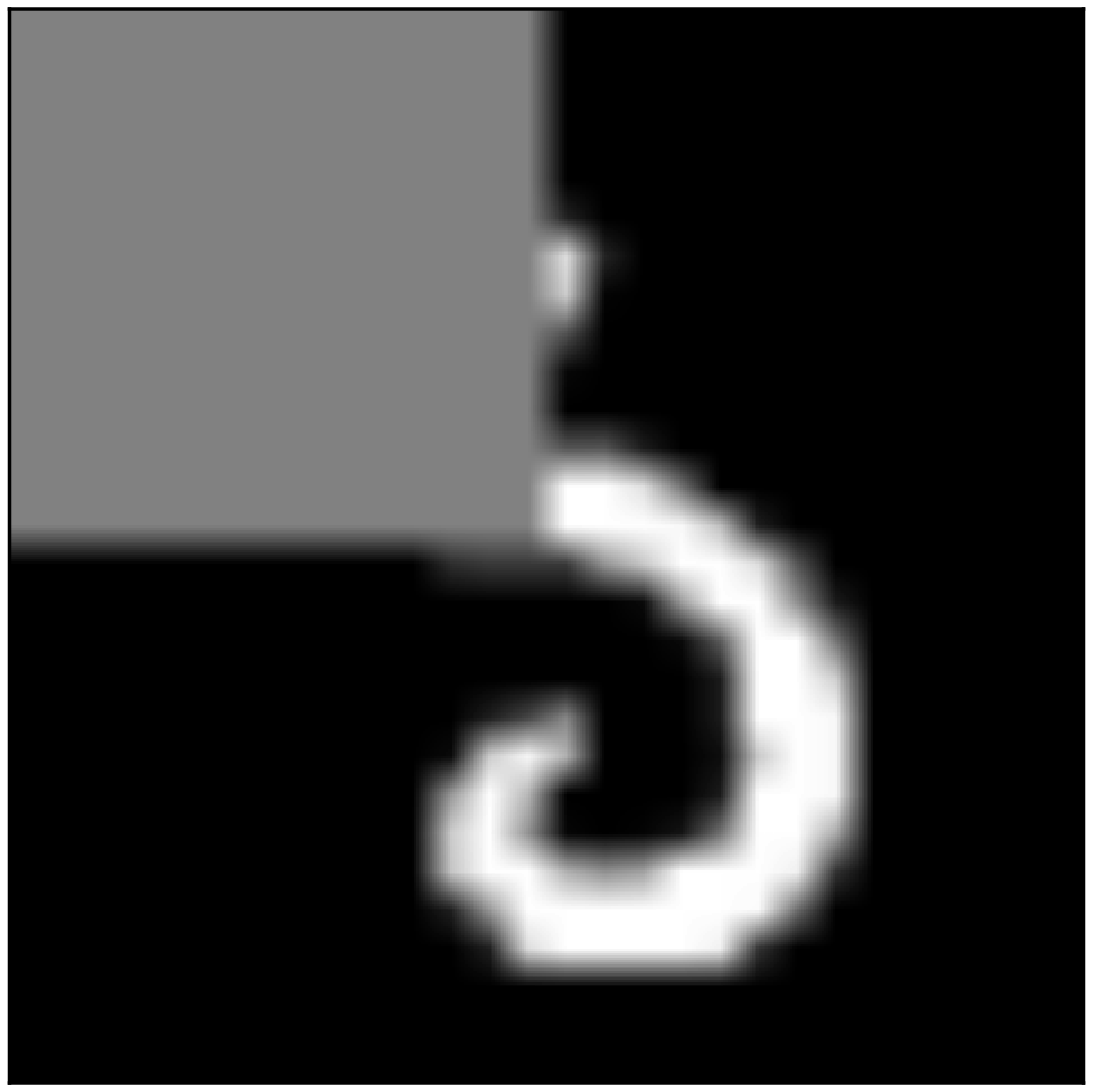}
        \includegraphics[width=0.12\textwidth]{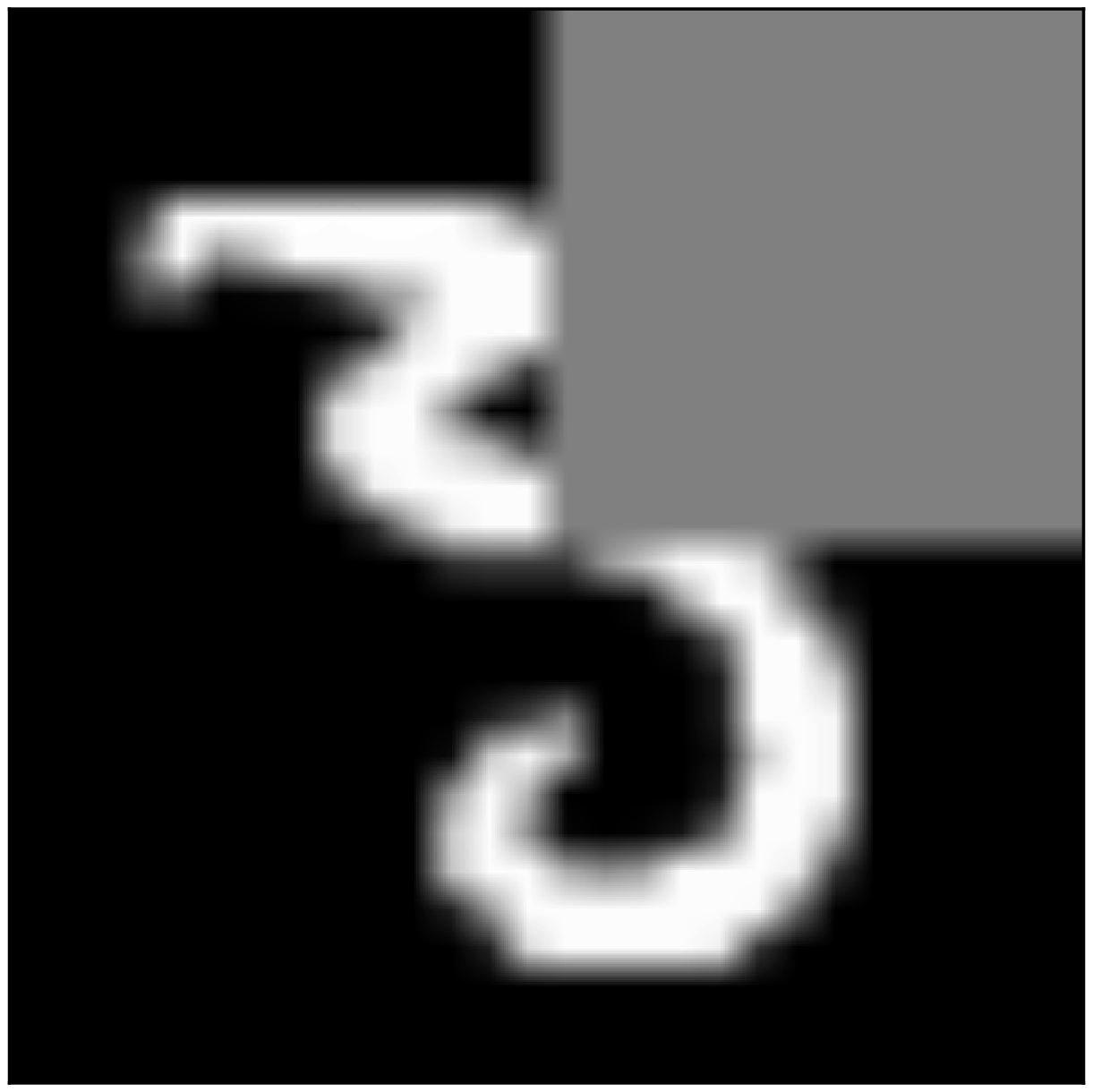}
        \includegraphics[width=0.12\textwidth]{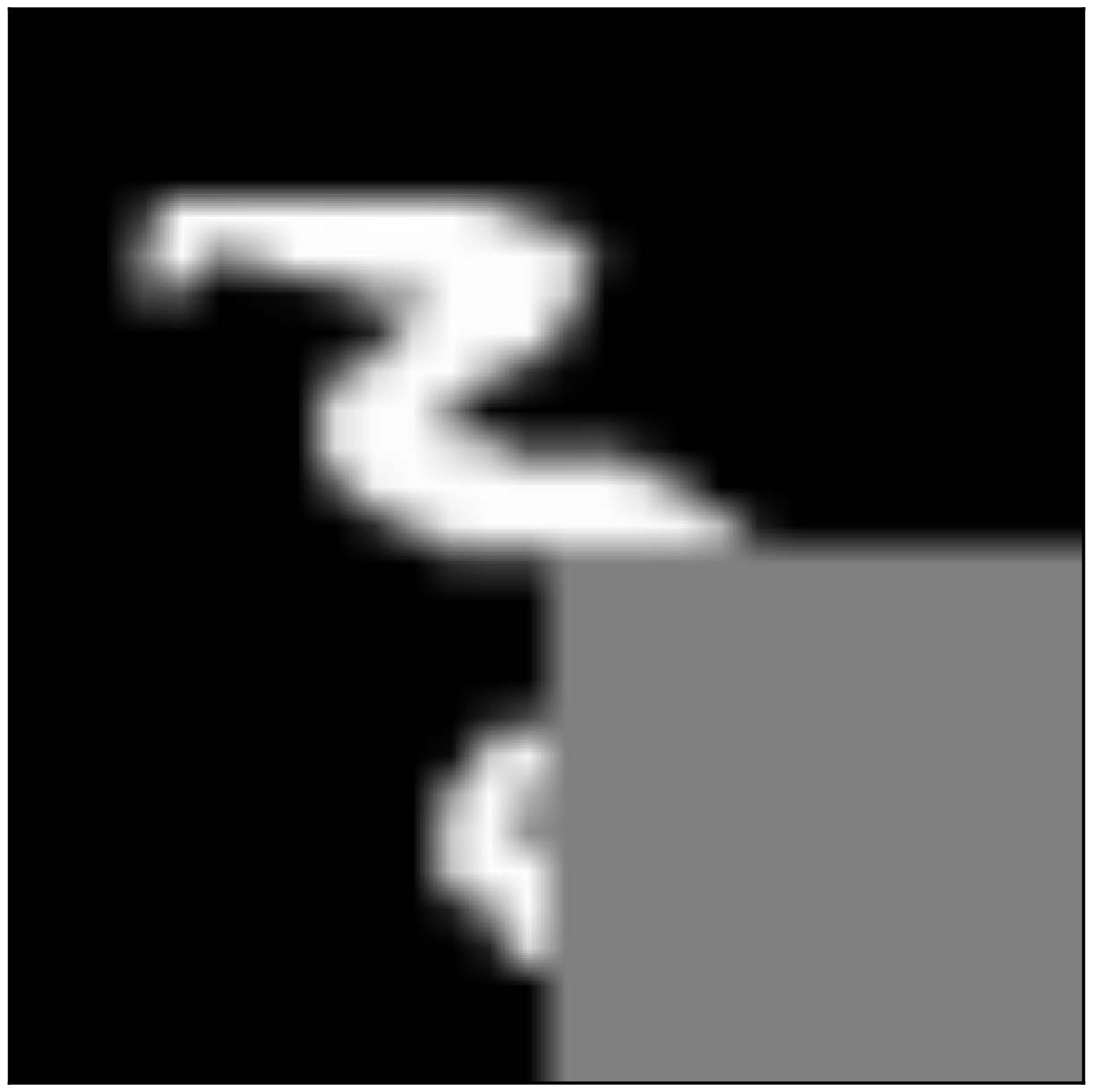}
        \includegraphics[width=0.12\textwidth]{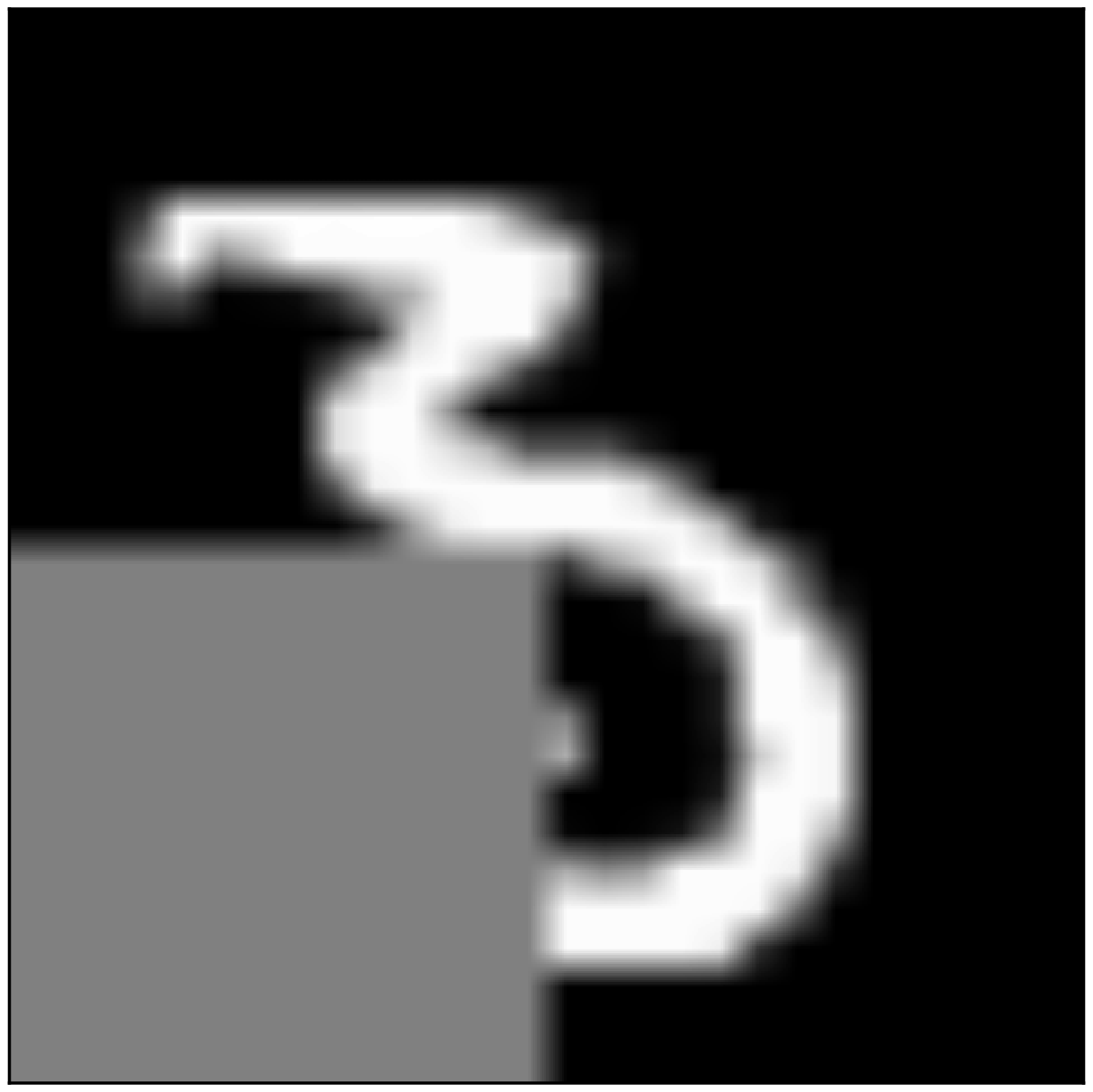}
        \label{fig:mnist_patterns}}
    \subfigure[]{
        \includegraphics[width=0.66\textwidth]{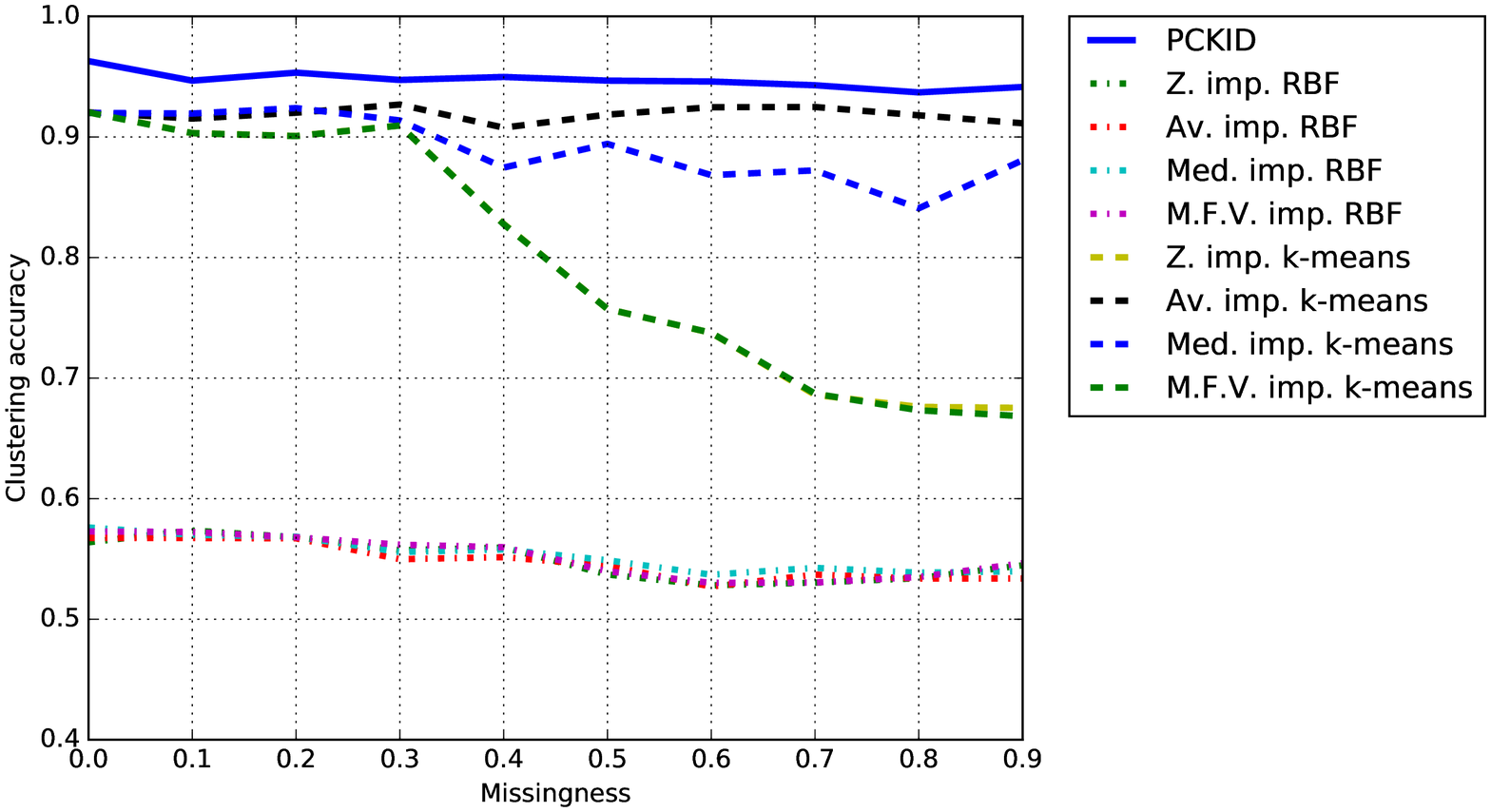}
        \label{fig:mnist_missing_56_accuracy}}
    \caption{(a): Example of missingness patterns. Gray pixels are considered missing. (b): Mean clustering accuracy as a function of the percentage of images with missing values.}
    \label{fig:mnist_patt_res}
\end{figure}
In this experiment, subsets containing 1000 of the MNIST 5 and 6 images are clustered.
The subsets consists of a balanced sample, i.e. there are approximately the same amount
    of images from each class.
The images are unraveled to 784 dimensional vectors, which are used as the input to
    the algorithms.
Missing data is generated by randomly choosing a proportion $p_\text{m}$ of the images and
    removing one of the four quadrants in the image according to the MAR mechanism.
These missingness patterns are illustrated in \figref{fig:mnist_patterns}.
In each test, we consider different probabilities of having missing quadrants, 
    i.e. $p_{\text{m}} \in \{0.0, 0.1, 0.2, \ldots, 0.9\}$,
Each method is run 30 times for each value of $p_{\text{m}}$, with a unique
    random subset of the data for each run.
Since there are dimensions in the dataset where there is no variation between images,
    they are removed before training the GMMs.
These are dimensions without information, and causes problems when inverting
    the covariance matrices.
The number of dimensions with variance varies across the runs, since the subset from
    the dataset and the missingness is randomly sampled for each run.
The number of dimensions with variance is approximately $500$.

\figref{fig:mnist_missing_56_accuracy} shows a plot of the mean clustering accuracy
    over the 30 runs versus the missingness proportion $p_\text{m}$. 
The proposed method outperforms the baseline methods for all $p_\text{m}$.
Although the clustering accuracy declines slightly when the $p_\text{m}$
    increases, the results are quite stable.
    
\begin{figure}[tbp]
    \centering
    \subfigure[]{
	    \includegraphics[width=0.49\textwidth]{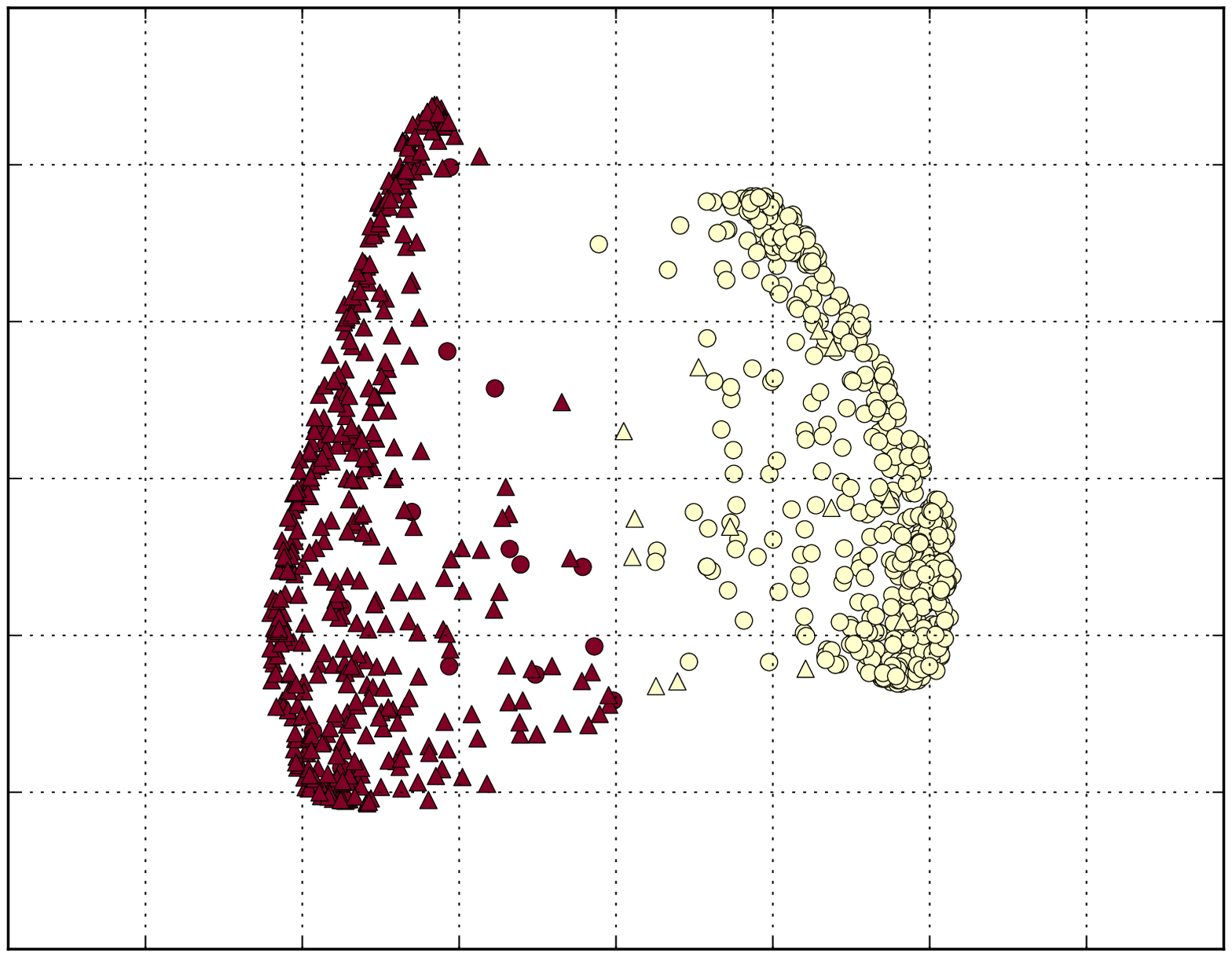}
	    \label{fig:mnist_map_00}}\hspace{-1.2em}
    \subfigure[]{
	    \scalebox{-1}[1]{
	    \includegraphics[width=0.49\textwidth]{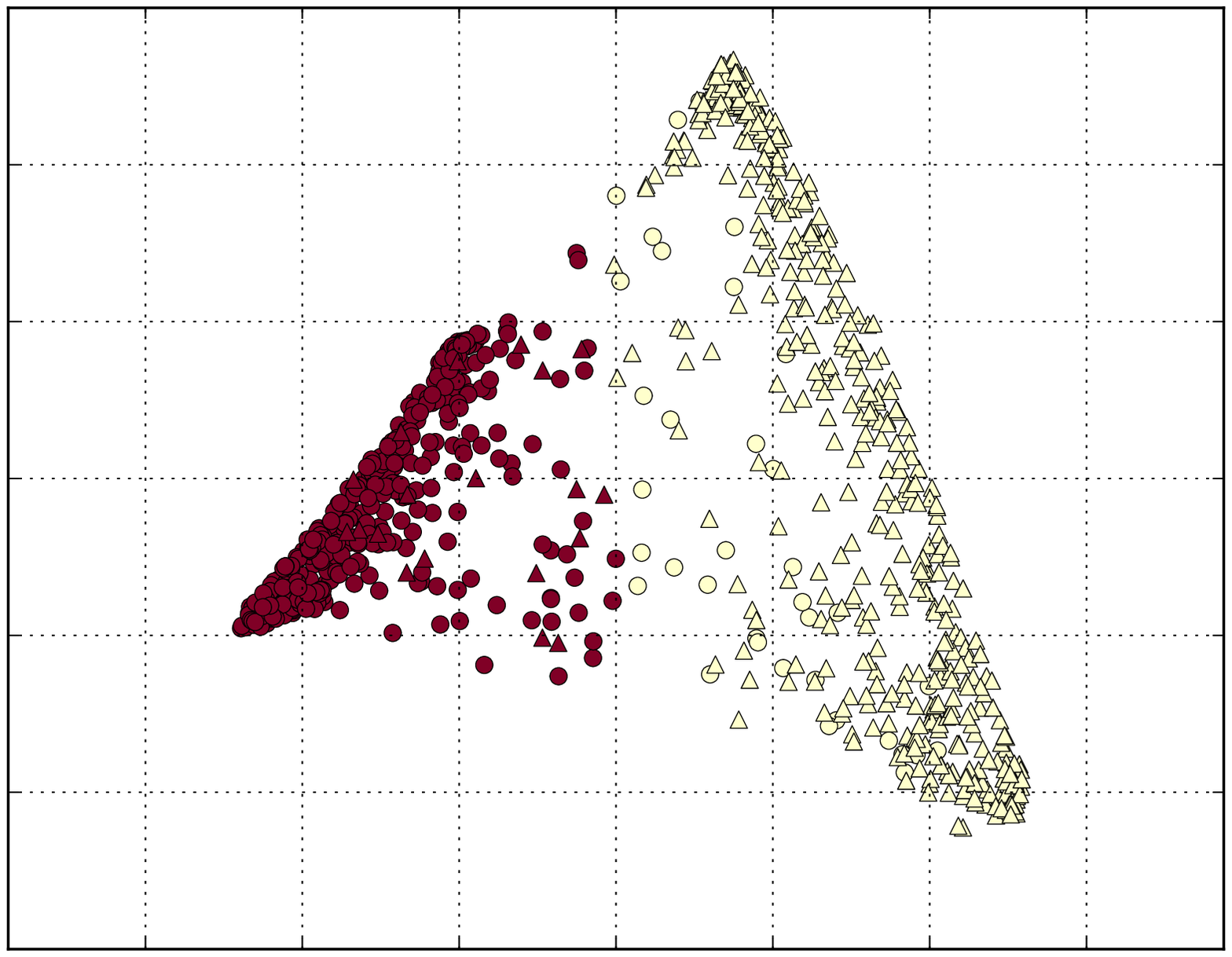}}
	    \label{fig:mnist_map_90}}\hspace{-1.2em}
    \caption{Example of embedding and clustering in kernel space with
    (a): No missingness, (b): 90\% missingness.
    The marker indicates the true label, while the color indicates the clustering results.}
    \label{fig:mnist_map}
\end{figure}

\figref{fig:mnist_map_00}--\figref{fig:mnist_map_90} shows two dimensional representations
    using kernel PCA on \PCKID{} with $p_\text{m} = 0$ and $p_\text{m} = 0.9$, respectively.
The shape of the markers indicate ground truth class, while the color indicate the clustering
    result.
It is interesting to see that although the plot with no missing data has a smoother structure,
    the overall topology seems to be very similar when $p_\text{m} = 0.9$.
The two classes seem to be less separable in the plot with more missing data,
    which is not surprising, given the numerical clustering results in
    \figref{fig:mnist_missing_56_accuracy}.

When considering the approach of $k$-means directly on data with imputed values,
    we see that none of the imputation techniques perform as well as \PCKID{},
    although in this case mean imputation works reasonably well.
To explain performance improvements as $p_\text{m}$ increases, it is possible that
    the missingness patterns chosen for this experiment introduce some noise
    that provides a form of regularization that is
    beneficial to certain imputation techniques, or maybe the balance in the
    dataset is helping the mean of the observed values to not introduce bias
    towards one class.
With median--, zero-- and most frequent value imputation, the clustering accuracy
    starts to decline around $p_\text{m} = 0.3$, with zero imputation and
    most frequent value imputation following almost exactly the same path.
This is likely due to the nature of the data, where many of the dimensions
    actually contains zeros in most of the images.
The most frequent value in most dimensions will then be zero.

Spectral clustering using an RBF kernel completely fails in this experiment, which
    is probably due to a sub-optimal kernel width.
However, this illustrates the difficulty with an unsupervised problem, where no
    prior information is given, making cross-validation virtually impossible
    without expertise knowledge on the data.

\subsection{Land cover clustering}
\bgroup
\def\arraystretch{1.05} 
\setlength\tabcolsep{.5em} 
\begin{table}[tbp]\scriptsize
  \centering
  \caption{\footnotesize Average clustering accuracy over 30 runs for different
          combinations of classes in the Hardangervidda dataset.
          The best results are marked in bold.
          The baseline methods are: ZI (zero imputation), AI (average imputation),
          MI (median imputation) and MFVI (most frequent value imputation), combined
          with either $k$-means or spectral clustering using an RBF kernel.}
  \label{tab:hardangervidda}
  \begin{tabular}{c|c|cccc|cccc}
     \cmidrule[1.5pt]{1-10}
\multirow{2}{*}{Classes} & \multirow{2}{*}{\PCKID{}} & \multicolumn{4}{c|}{Spectral clustering, RBF} & \multicolumn{4}{c}{$k$-means} \\
&& ZI & AI & MI & MFVI & ZI & AI & MI  & MFVI \\
     \cmidrule[1.5pt]{1-10}
2-3 & 0.580          & 0.610 & 0.610 & 0.624 & \textbf{0.627} & 0.601          & 0.601          & 0.601 & 0.605 \\
2-4 & 0.536          & 0.663 & 0.663 & 0.663 & \textbf{0.674} & 0.591          & 0.591          & 0.590 & 0.597 \\
2-5 & 0.661          & 0.589 & 0.589 & 0.598 & 0.605          & \textbf{0.671} & \textbf{0.671} & 0.663 & 0.652 \\
2-6 & \textbf{0.712} & 0.578 & 0.578 & 0.571 & 0.594          & 0.672          & 0.672          & 0.664 & 0.639 \\
2-7 & \textbf{0.868} & 0.519 & 0.519 & 0.516 & 0.501          & 0.854          & 0.854          & 0.858 & 0.862 \\
3-4 & 0.698          & 0.505 & 0.505 & 0.505 & 0.511          & 0.697          & 0.697          & 0.711 & \textbf{0.722} \\
3-5 & \textbf{0.563} & 0.521 & 0.521 & 0.511 & 0.516          & 0.534          & 0.534          & 0.540 & 0.540 \\
3-6 & \textbf{0.620} & 0.565 & 0.565 & 0.562 & 0.564          & 0.521          & 0.521          & 0.519 & 0.523 \\
3-7 & \textbf{0.933} & 0.501 & 0.501 & 0.726 & 0.522          & 0.577          & 0.577          & 0.599 & 0.603 \\
4-5 & 0.764          & 0.517 & 0.517 & 0.512 & 0.510          & 0.839          & 0.839          & 0.847 & \textbf{0.848} \\
4-6 & \textbf{0.897} & 0.517 & 0.517 & 0.547 & 0.547          & \textbf{0.897} & \textbf{0.897} & 0.894 & 0.880 \\
4-7 & \textbf{0.931} & 0.550 & 0.550 & 0.547 & 0.534          & 0.687          & 0.687          & 0.687 & 0.718 \\
5-6 & \textbf{0.740} & 0.623 & 0.623 & 0.644 & 0.672          & 0.554          & 0.554          & 0.602 & 0.606 \\
5-7 & \textbf{0.956} & 0.687 & 0.687 & 0.667 & 0.698          & 0.706          & 0.706          & 0.706 & 0.706 \\
6-7 & \textbf{0.970} & 0.767 & 0.767 & 0.752 & 0.696          & 0.759          & 0.759          & 0.759 & 0.670 \\
     \cmidrule[1.5pt]{1-10}
  \end{tabular}
\end{table}
\egroup
In this experiment, we cluster pixels in high resolution land cover images
    contaminated with clouds, also used for classification in
    \cite{salberg2011land, salberg2012land}.
The data consists of three Landsat ETM+ images covering Hardangervidda in
    southern Norway, in addition to elevation and slope information.
With 6 bands in each image, the total dimensionality of the data is 20.
In this dataset, a value is considered missing if a pixel in an
    image is contaminated by either clouds or snow/ice.
For details on how the dataset is constructed, see \cite{salberg2011land}.

The pixels in the image are labeled as one of 7 classes: 1) \textit{water},
    2) \textit{ridge}, 3) \textit{leeside}, 4) \textit{snowbed},
    5) \textit{mire}, 6) \textit{forest} and 7) \textit{rock}.
In this experiment, we exclude the water class, since it is easy to separate
    from the other classes in the Norwegian mountain vegetation.
To investigate how the \PCKID{} handle the different combination of classes,
    we restrict the analysis to pairwise classes.
Each dimension is standardized on the observed data.

The average clustering accuracy for each combination of the chosen classes
    is reported in \tabref{tab:hardangervidda}.
The average is computed over 30 runs of each algorithm.
We see that \PCKID{} seems to perform better for most class pairs.
Although it might struggle with some classes, most notably class 2.
For the class pair 3-5, \PCKID{} wins with a clustering accuracy of 0.563,
    which is not much better than random chance in a two class problem.
It is however worth to note that the classes labels are set according the
    vegetation at the actual location, which is not necessarily the group
    structure reflected in the data.
The class combinations where \PCKID{} really outperforms the other
    methods seems to be when class 7 (rocks) is present in the data,
    where we improve performance by up to 25 percentage points 
    with regards to the baseline methods.
\begin{SCfigure}[1.2][tbp]
    \centering
    \includegraphics[width=0.46\textwidth,trim={0.0cm 0.0cm 0.0cm 0.0cm},clip]
    {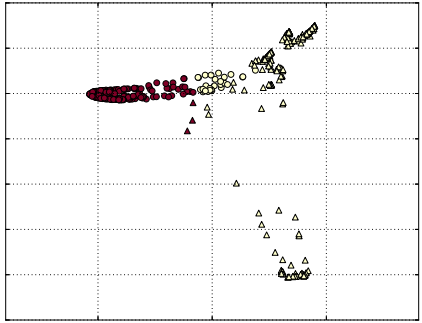}
    \caption{Example of mapping for the \textit{forest}--\textit{rock} class pair. Colors indicate clustering,
    while the shape of the marker indicate the ground truth label.}
    \label{fig:hardangervidda_cluster_mapping}
\end{SCfigure}

\figref{fig:hardangervidda_cluster_mapping} shows an example of a mapping
    for the \textit{forest}--\textit{rock} class pair, where it seems like the \textit{rock}
    class, as defined by the ground truth, actually consists of two separate structures
    in the KPCA embedding using \PCKID{}.
This demonstrates the power of \PCKID{}s ability to adapt to the inherent structures
    in the data.

\section{Conclusion}  
In this paper, we have proposed \PCKID{}, a novel kernel function for spectral
    clustering, designed to
    i) explicitly handle incomplete data and
    ii) be robust with regards to parameter choice.
By combining posterior distributions of Gaussian Mixture Models for incomplete
    data on different scales, \PCKID{} is able to learn similarities on the
    data manifold, yielding a kernel function without any \textit{critical}
    hyperparameters to tune.
Experiments have demonstrated the strength of our method, by improved clustering
    accuracy compared to baseline methods, while keeping parameters fixed
    for all experiments.

\subsubsection*{Acknowledgments}
This work was partially funded by the Norwegian Research Council FRIPRO grant no.\ 239844 on developing the \emph{Next Generation Learning Machines}.

\bibliographystyle{splncs03}
\bibliography{references.bib}
\clearpage

\end{document}